# A Flexible Job Shop Scheduling Problem Involving Reconfigurable Machine Tools under Industry 5.0

Hessam Bakhshi-Khaniki[1[0009-0003-3251-9878]], Reza Tavakkoli-Moghaddam[1[0000-0002-6757-926X]] (✉), Zdenek Hanzalek[2[0000-0002-8135-1296]], and Behdin Vahedi-Nouri[1[0000-0002-6568-6782]]

[1] School of Industrial Engineering, College of Engineering, University of Tehran, Tehran, Iran
[2] Industrial Informatics Department, Czech Institute of Informatics Robotics and Cybernetics, Czech Technical University in Prague, Prague, Czech Republic
tavakoli@ut.ac.ir

**Abstract.** The rise of Industry 5.0 has introduced new demands for manufacturing companies, requiring a shift in how production schedules are managed to address human-centered, environmental, and economic goals comprehensively. The flexible job shop scheduling problem (FJSSP), which involves processing operations on various capable machines, accurately reflects the complexities of modern manufacturing settings. This paper investigates the FJSSP involving reconfigurable machine tools with configuration-dependent setup times, while integrating human aspects like worker assignments, moving time, and rest periods, as well as minimizing total energy consumption. A mixed-integer programming (MIP) model is developed to simultaneously optimize these objectives. The model determines the assignment of operations to machines, workers, and configurations while sequencing operations, scheduling worker movements, and respecting rest periods, and minimizing overall energy consumption. Given the NP-hard nature of the FJSSP with worker assignments and reconfigurable tools, a memetic algorithm (MA) is proposed. This meta-heuristic evolutionary algorithm features a three-layer chromosome encoding method, specialized crossover and mutation strategies, and neighborhood search mechanisms to enhance solution quality and diversity. Comparisons of MA with MIP and genetic algorithms (GA) on benchmark instances demonstrate the MA's efficiency and effectiveness, particularly for larger problem instances where MIP becomes impractical. This research paves the way for sustainable and resilient production schedules tailored for the factory of the future under the Industry 5.0 paradigm. The work bridges a crucial gap in current literature by integrating worker and environmental impact into the FJSSP with reconfigurable machine models.

## 1 Introduction

As manufacturing enterprises transition from the Industry 4.0 era towards the emerging Industry 5.0 paradigm, they are faced with new imperatives around human-centricity, environmental sustainability, and operational resilience. Scheduling, which lies at the heart of production operations, can no longer purely focus on economic objectives like minimizing makespan and costs. There is a critical need to holistically account for human factors related to worker well-being and social sustainability, as well as environmental impact factors like energy consumption and emissions.

The flexible job shop scheduling problem (FJSSP), in which operations can be processed on multiple capable machines, accurately represents the modern production environment [1]. Integrating human aspects (e.g., worker skills, ergonomics, fatigue management, and the human dimension) into the FJSSP is essential for enabling human-centric manufacturing. Simultaneously, it is vital to incorporate environmental criteria like energy efficiency, emissions reduction, and waste minimization to drive sustainable production schedules. The shift towards Industry 5.0 necessitates manufacturing companies to redefine their priorities, placing equal emphasis on economic, human, and environmental criteria. By holistically incorporating these multi-faceted objectives into scheduling decisions, companies can drive operational excellence while fostering workforce prosperity and environmental stewardship. The survey paper [2] underscores vital research opportunities to develop scheduling techniques that harmoniously balance these key dimensions for the factory of the future.

The well-being of workers is optimized through ergonomic scheduling, rest periods, and safe conditions. Skills development includes versatile scheduling and on-the-job learning. Human-machine collaboration enhances tasks, while decentralized decision-making empowers workers. Social sustainability fosters teamwork and work-life balance. Together, these practices prioritize workers' needs and skills, promoting adaptability and resilience [3].

The job shop scheduling problem (JSSP) has extensively been studied over recent decades, with researchers proposing various approaches to support scheduling decisions in evolving production environments. The FJSSP, which allows operations to be processed on multiple capable machines, better represents real-world manufacturing systems and has gained significant attention. Traditional JSSP and FJSSP research has predominantly focused on optimizing economic objectives like minimizing makespan, production costs, and tardiness. However, with the advent of Industry 4.0 and now Industry 5.0, new priorities around human-centricity, sustainability, and resilience have emerged [4].

Many researchers are exploring FJSSP models that incorporate human factors, such as worker skills [4], ergonomic risks [5], fatigue management [6], and labor costs [7]. He et al. [8] developed an improved African vulture optimization algorithm to solve the dual-resource constrained FJSSP, which minimizes makespan and total delay. Tan et al. [6] proposed an enhanced NSGA-II for solving a fatigue-conscious dual resource-constrained FJSSP, which minimizes worker fatigue and makespan by jointly scheduling machines and workers, which has practical relevance in flexible manufacturing systems. Müller et al. [4] addressed an FJSSP with workforce constraints to minimize makespan. It considers a heterogeneous workforce, in which workers may possess different qualifications for operating machines, affecting processing times. They proposed filter-and-fan heuristic approaches that combine local search with tree search, decom-posing the problem into machine allocation/sequencing and worker assignment decisions. Hongyu et al. [5] proposed an enhanced survival duration-guided NSGA-III for a sustainable FJSSP with dual resources, aiming to optimize makespan, energy consumption, and ergonomic risk while scheduling machines and workers.

In parallel, prompted by global environmental concerns, researchers have investigated integrating objectives and constraints related to energy consumption [9], carbon emissions [10], and noise pollution [11] into green FJSSP formulations. Jia et al. [9] developed an enhanced GA utilizing a four-layer chromosome encoding scheme and improved initialization, crossover, and mutation operations for solving the green FJSSP that considers cost, carbon emissions, and customer satisfaction under the time of use electricity pricing. Wang et al. [10] addressed concerns regarding the low-carbon, many-objective FJSSP in the context of global warming and energy crises. It aims to minimize completion time, total delay time, processing load rate of the bottleneck machine, and total carbon emissions of the system. Hajibabaei et al. [11] introduced a mathematical model and a Lagrangian relaxation algorithm for an FJSSP incorporating an assembly stage, machine breakdowns, and batch transportation to customers. The model addresses tardiness/earliness costs, fuzzy transportation/makespan costs, fuzzy $CO_2$ emissions, and noise pollution resulting from increased machine speeds. However, simultaneous optimization of economic, environmental, and human factors remains largely unexplored.

In recent years, reconfigurable manufacturing systems and reconfigurable machine tools (RMTs) have emerged as a new manufacturing paradigm to meet rapidly changing production requirements. The modular structure of RMTs enables easy reconfiguration for performing different operations or changing their functional capabilities. This reconfigurability allows RMTs to process various operation types by changing tool attachments, adding/removing functional modules, etc. [12].

Scheduling problems involving RMTs pose new challenges compared to the classical FJSSP. In addition to the normal decisions of operation-to-machine assignment and operation sequencing, scheduling with RMTs requires determining the configuration of each machine for processing its assigned operations. Furthermore, setup times for switching a machine's configuration are often sequence-dependent. Fan et al. [13] proposed an enhanced GA for the FJSSP with machine reconfigurations, in which machines adapt with auxiliary modules for various tasks. Fan et al. [14] investigated the FJSSP that considers lot-streaming and machine reconfigurations to minimize total weighted tardiness. To address this problem involving both discrete and continuous optimization attributes, they proposed a matheuristic algorithm that combines a GA framework with two mixed integer linear programming-based local search strategies for optimizing lot-sizing plans. Guo et al. [15] proposed a novel dynamic FJSSP with reconfigurable manufacturing cells, wherein each cell is capable of executing various operations, requiring reconfiguration time during transitions between operations. The authors devised an enhanced genetic programming hyper-heuristic method, integrating an individual simplification policy to address the FJSSP, accounting for objectives like completion time, delay time, and reconfiguration time.

The existing body of research concerning integrated sustainable FJSSP models is notably scarce when it comes to considering both worker well-being and environmental impact concurrently. This gap in the literature is particularly pronounced in the realm of reconfigurable machine tools. Reconfigurable machine tools, with their ability to adapt to various manufacturing tasks and environments, offer a promising avenue for enhancing sustainability in manufacturing processes. However, the dearth of studies addressing their role within integrated sustainable FJSSP models represents a significant lacuna in current research efforts. Addressing this gap could yield valuable insights into optimizing manufacturing processes not only in terms of efficiency but also in promoting worker welfare and minimizing environmental harm.

## 2 Problem Description

The FJSSP, in which machines are reconfigurable with configuration-dependent setup times, considers both human and environmental factors, specifically minimizing total energy consumption while limiting worker assignment. This problem involves sets of jobs $J$, machines $K$, workers $L$, and configurations $C_K$. Each job requires a sequence of operations that need to be processed with a predefined sequence and each operation can be completed in configurations by a subset of workers and machines. The assumptions and limitations of our model are as follows:

- Jobs are independent.
- There is a set of reconfigurable machines on the shop floor, each with multiple configurations.
- Each operation can be processed on at least one configuration of one of the reconfigurable machines.
- Setup time between changing configurations is considered.
- Each machine can be in only one configuration and process one operation at a time.
- Preemption of operations is not allowed.
- Both human (i.e., workers) and machine resources are considered.
- Worker moving time is considered when they need to move to another machine.
- Workers rest after each operation is completed.
- Workers rest for up to 10% of their work time.

The notations used in the proposed mathematical model are summarized as follows:

**Indices:**

| | |
|---|---|
| $i, i'$ | Jobs index |
| $j, j'$ | Operations index |
| $k, k'$ | Machines index |
| $l$ | Worker index |
| $c, c_1, c_2$ | Configuration index |

**Sets:**

| | |
|---|---|
| $J$ | Set of jobs |
| $K$ | Set of machines |

$L$ Set of workers

$C_k$ Set of configurations of machine $k$

**Parameters:**

$N$ Total number of jobs

$M$ Total number of machines

$L$ Total number of workers

$n_i$ Total number of operations of job $i$

$U$ A big positive number

$N$ Total number of jobs

$O_{ij}$ Operation $j$ of job $i$

$M_{ij}$ Subsets of machines for $O_{ij}$

$W_k$ Subsets of workers for machine $k$

$Pt_{ij}$ Processing time of $O_{ij}$

$Se_{c1,c2,k}$ Setup time of changing configuration from $C_1$ to $C_2$ on machine $k$

$Mt_{kk'}$ Worker moving time between machine $k$ and $k'$

$Rt_{ijl}$ Rest time of worker $l$ after performing operation $O_{ij}$

$E_{ijkl}$ Energy consumption of operation $O_{ij}$ on machine $k$ performed by worker $l$

$AE$ Auxiliary energy

$r_{ijklc}$ 1 if operation $O_{ij}$ can be processed on machine $k$ on configuration $c$ by worker $l$; 0, otherwise

**Decision variables:**

$TE$ Total energy consumption

$C_{max}$ Makespan

$C_{ij}$ Completion time of $O_{ij}$

$St_{ij}$ Starting time of processing of $O_{ij}$

$X_{ijklc}$ 1 if machine $k$ and worker $l$ on configuration $c$ are selected for operation $O_{ij}$; 0, otherwise

$Y_{iji'j'}$ 1 if operation $O_{ij}$ is performed before operation $O_{i'j'}$; 0, otherwise

$Z_{i'j'kk'l}$ 1 if worker $l$ moves from machine $k$ to perform operation $O_{i'j'}$ on machine $k'$; 0, otherwise

The mathematical model is defined as follows:

$$Min\ TE = \mathrm{AE}C_{max} + \sum_{i \in I} \sum_{j \in J} \sum_{k \in K} \sum_{c \in C_k} \sum_{l \in W_k} E_{ijkl} Pt_{ijklc} X_{ijklc} \quad (1)$$

s.t.

$$C_{ij} = St_{ij} + Pt_{ijklc}X_{ijklc} \quad \forall i \in J, j \in n_i, k \in K, l \in L, c \in C_K \tag{2}$$

$$C_{max} \geq C_{in_i} \quad \forall i \in J \tag{3}$$

$$\sum_{k \in M_{ij}} \sum_{c \in C_k} \sum_{l \in W_k} X_{ijklc} = 1 \quad \forall i \in J, j \in O_{ij} \tag{4}$$

$$X_{ijklc} \leq r_{ijklc} \quad \forall i \in J, j \in O_{ij}, k \in M_{ij}, l \in W_K, c \in C_K \tag{5}$$

$$St_{ij} - St_{i,j-1} \geq \sum_{k \in M_{ij}} \sum_{c \in C_k} \sum_{l \in W_k} Pt_{ijklc}X_{ijklc} \quad \forall i \in J, j \in O_{ij} \tag{6}$$

$$\begin{gathered} St_{i'j'} - St_{ij} \geq Pt_{ijklc}X_{ijklc} + Mt_{kk'}Z_{i'j'kk'l} + Rt_{ijl}\left(1 - Z_{i'j'kk'l}\right) \\ \forall i, i' \in J, j \in O_{ij}, j' \in O_{i'j'}, O_{ij} \neq O_{i'j'}, k, k' \in M_{ij}, l \in W_K, c \in C_K \end{gathered} \tag{7}$$

$$\begin{gathered} St_{ij} - St_{i'j'} \geq Pt_{ijklc_2} + Se_{c_1,c_2,k} - \left(2 - X_{ijklc_2} - X_{i'j'klc_1} + Y_{iji'j'}\right)U \\ \forall i, i' \in J, j \in O_{ij}, j' \in O_{i'j'}, O_{ij} \neq O_{i'j'}, k \in M_{ij}, l \in W_K, c_1, c_2 \in C_K \end{gathered} \tag{8}$$

$$\begin{gathered} St_{i'j'} - St_{ij} \geq Pt_{i'j'klc_2} + Se_{c_1,c_2,k} - \left(3 - X_{ijklc_1} - X_{i'j'klc_2} - Y_{iji'j'}\right)U \\ \forall i, i' \in J, j \in O_{ij}, j' \in O_{i'j'}, O_{ij} \neq O_{i'j'}, k \in M_{ij}, l \in W_K, c_1, c_2 \in C_K \end{gathered} \tag{9}$$

$$St_{ij}, C_{ij}, C_{max} \geq 0 \quad \forall i \in J, j \in O_{ij} \tag{10}$$

$$X_{ijklc}, Y_{iji'j'}, Z_{i'j'kk'l} \in \{0,1\} \quad \forall i \in J, j \in O_{ij}, k \in M_{ij}, l \in W_K, c \in C_K \tag{11}$$

Equation (1) represents the objective function, aiming to minimize the total energy consumption. Equation (2) calculates the completion time of each operation based on the starting time and processing time. Constraint (3) is used to determine makespan, while constraint (4) ensures that each operation is processed by one machine and one worker in the one-machine configuration. Constraint (5) guarantees that operations can be processed only on machine configurations, for which they are eligible. Constraint (6) ensures that each operation of the same job should not be processed before the previous operation is completed. Constraint (7) determines the starting time of an operation considering worker rest time and moving time, while Constraints (8) and (9) prevent operations from overlapping. Lastly, Constraints (10) and (11) define the decision variables.

# 3 Methodology

The FJSP with worker assignment and machine reconfigurable tools is an NP-hard combinatorial optimization problem, making it challenging to find optimal solutions for medium and large instances in a short time. As a result, obtaining optimal solutions using exact methods within a reasonable timeframe is extremely difficult. To address this issue, memetic algorithm (MA), which is a meta-heuristic evolutionary algorithm, is well-suited. The key steps of the developed MA are as follows:

**Step 1:** Set parameters.

**Step 2:** Chromosome encoding and decoding.

**Step 3:** Initialize population. Set the initial generation count $i = 0$ and initialize population P(i) with n individuals.

**Step 4:** Generate offspring population $F_1(i)$ using the crossover method.

**Step 5:** Generate offspring population $F_2(i)$ using the mutation method.

**Step 6:** Generate population $Q(i)$ by combining $F_1$(i), $F_2(i)$, and $P(i)$.

**Step 7:** Perform neighborhood strategies $N_1$, $N_2$, and $N_3$ to improve the quality of individuals in $Q(i)$ and set $P(i+1)= Q(i)$.

**Step 8:** If the termination criteria are not satisfied, return to Step 4; otherwise, output $P(i+1)$.

The FJSSP with machine reconfigurable tools employs a three-layer chromosome encoding method, as illustrated in Fig. 1.

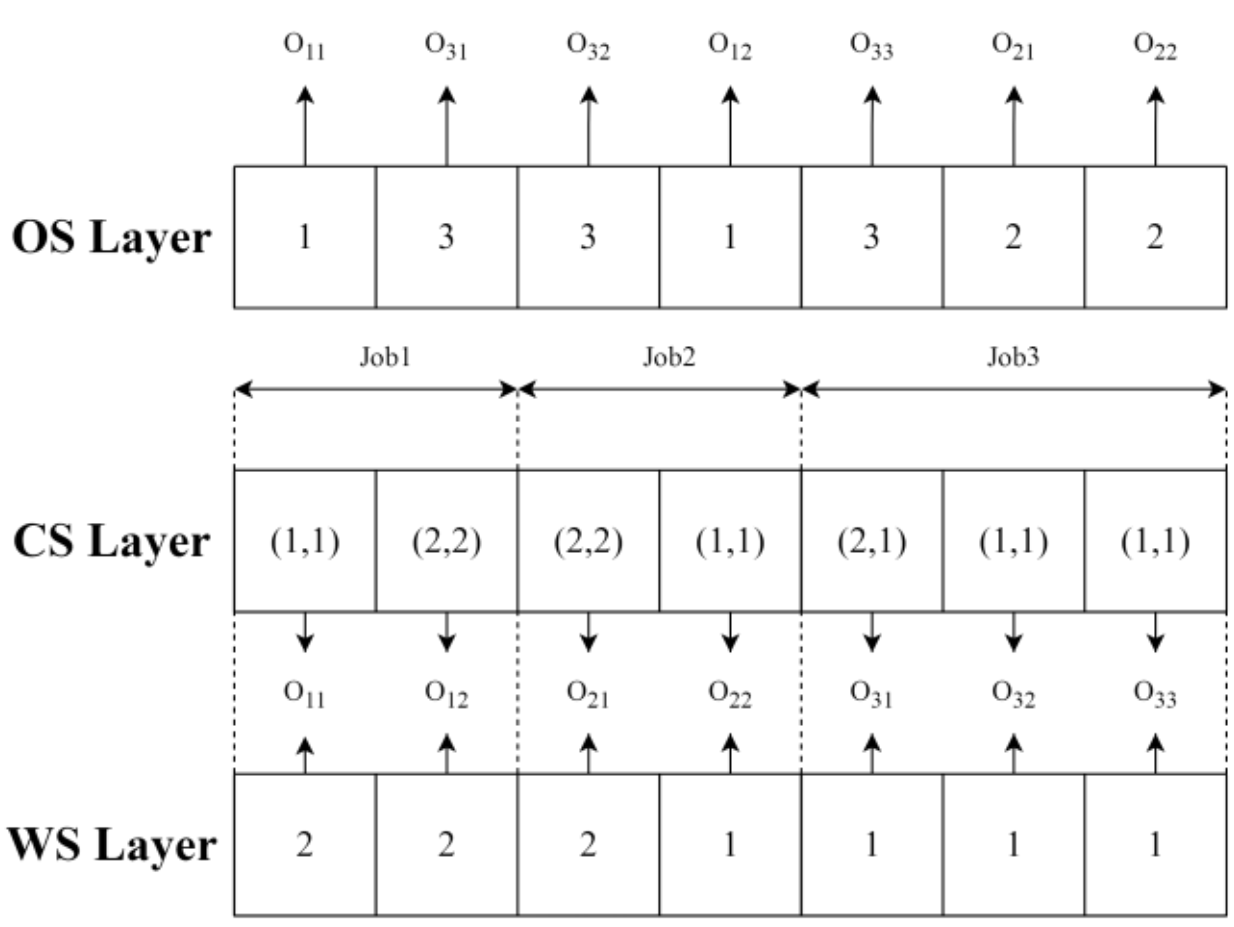

**Fig. 1.** Encoding instance.

In this figure, the three layers are as follows:

**Operation Sequencing (OS) Layer**: Represented as [1, 3, 3, 1, 3, 2, 2], where each element corresponds to operations $O_{11}$, $O_{31}$, $O_{32}$, $O_{12}$, $O_{33}$, $O_{21}$ and $O_{22}$, respectively.

**Configuration Selection (CS) Layer**: This layer includes both machine assignment and machine configuration. It is represented as [(1,1), (2,2), (2,2), (1,1), (2,1), (1,1), (1,1)]. Here, the first element denotes the machine, and the second element denotes its configuration, corresponding to operations $O_{11}$, $O_{12}$, $O_{21}$, $O_{22}$, $O_{31}$, $O_{32}$, and $O_{33}$, respectively.

**Worker Selection (WS) Layer:** Represented as [2, 2, 2, 1, 1, 1, 1], where each element indicates the worker assigned to the respective operations $O_{11}$, $O_{12}$, $O_{21}$, $O_{22}$, $O_{31}$, $O_{32}$, and $O_{33}$.

For the initial population, the smallest energy consumption (SEC) method is used to design the OS, CS, and WS layers. In this method, workers, machines, and configurations are assigned based on the smallest energy consumption for each operation.

The crossover generator is employed to improve search efficiency. In this paper, we utilize a random selection process, where parents (i.e., $P_1$ and $P_2$) produce offspring (i.e., $F_1$ and $F_2$) as shown in Fig. 2(a). We develop a job-based crossover (JBX) mechanism specifically for the OS layer. This method involves creating two distinct sets of jobs. For offspring $F_1$, the job set from parent $P_1$ is directly assigned to $F_1$, preserving the exact positions of the jobs. The remaining positions in $F_1$ are filled with jobs from parent $P_2$ that are not present in the job set of $P_1$, maintaining the order in which these jobs appear in $P_2$. Similarly, for offspring $F_2$, the job set from parent $P_2$ is directly assigned to $F_2$, maintaining the exact positions of these jobs. The remaining positions in $F_2$ are filled with jobs from parent $P_1$ that are not included in the job set of $P_2$, preserving their original order from $P_1$. This method ensures that the offspring inherit a mix of job assignments from both parents while maintaining the structure and integrity of job positions from each parent's respective job set. By doing so, the JBX method enhances the diversity and quality of the solutions generated in the search process, potentially leading to better optimization outcomes.

Furthermore, a multi-configuration crossover (MCX) and a multi-worker crossover (MWX) are developed for the CS and WS layers as shown in Figs. 2(b) and 2(c), respectively. This process begins with the generation of a random chromosome composed of binary values (i.e., 0 and 1). For offspring $F_1$, genes located at positions marked by a value of 0 are directly inherited from parent $P_1$. Conversely, genes at positions marked by a value of 1 are swapped with the corresponding genes in parent $P_2$. In parallel, for Offspring $F_2$, positions marked by a value of 0 inherit genes from parent $P_2$, while positions marked by a value of 1 are exchanged with the corresponding genes in parent $P_1$.

To perform mutations in our algorithm, we employ different strategies for each layer. In the OS layer, we utilize the swapping method, which involves selecting genes at random and exchanging their positions. For instance, in Fig. 3(a), genes at positions [2, 5] are swapped with each other. In the CS layer, we randomly select several genes and replaced them with eligible machines and configurations. As depicted in Fig. 3(b), genes at positions [2, 5, 7] are substituted with new, eligible machine and configuration pairs. Similarly, in the WS layer, multiple genes are randomly selected and replaced with eligible workers. As demonstrated in Fig. 3(c), genes at positions [2, 4, 5] are replaced with eligible workers.

To improve our algorithm with neighborhood search, a method is integrated that systematically explores the neighborhood of a solution to uncover enhancements. Three strategies for neighborhood search are devised as follows:

**N1**: In this strategy, the operations of two different jobs are randomly exchanged. Then, the corresponding genes in the CS layer at the same positions are selected. Subsequently, the machines and configurations for these genes are altered accordingly, and an eligible worker is selected.

**N2**: This strategy entails two genes being selected in the OS layer, and the order of the genes between them being reversed. Next, the corresponding genes in the CS layer at the same positions are selected, their machines and configurations are adjusted, and then an eligible worker is selected.

**N3**: In this approach, two genes are randomly selected in the OS layer. Then, the position of the second gene is moved to immediately follow the position of the first gene, shifting all intervening genes to the right. Following this adjustment, the corresponding genes in the CS layer at the same positions are selected, their machines and configurations are modified, and then an eligible worker is selected.

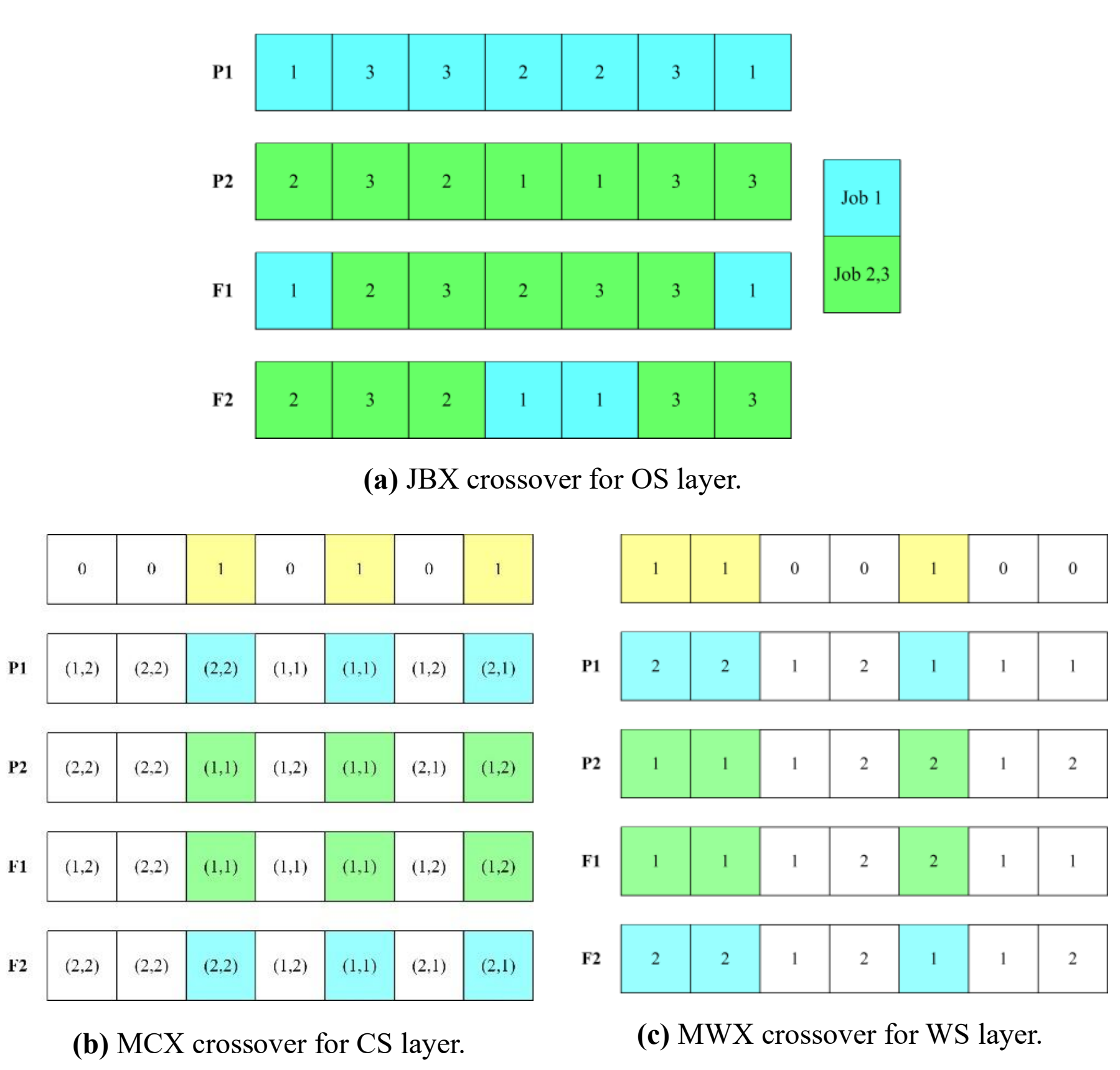

**(a)** JBX crossover for OS layer.

**(b)** MCX crossover for CS layer.

**(c)** MWX crossover for WS layer.

**Fig. 2.** Crossover generator.

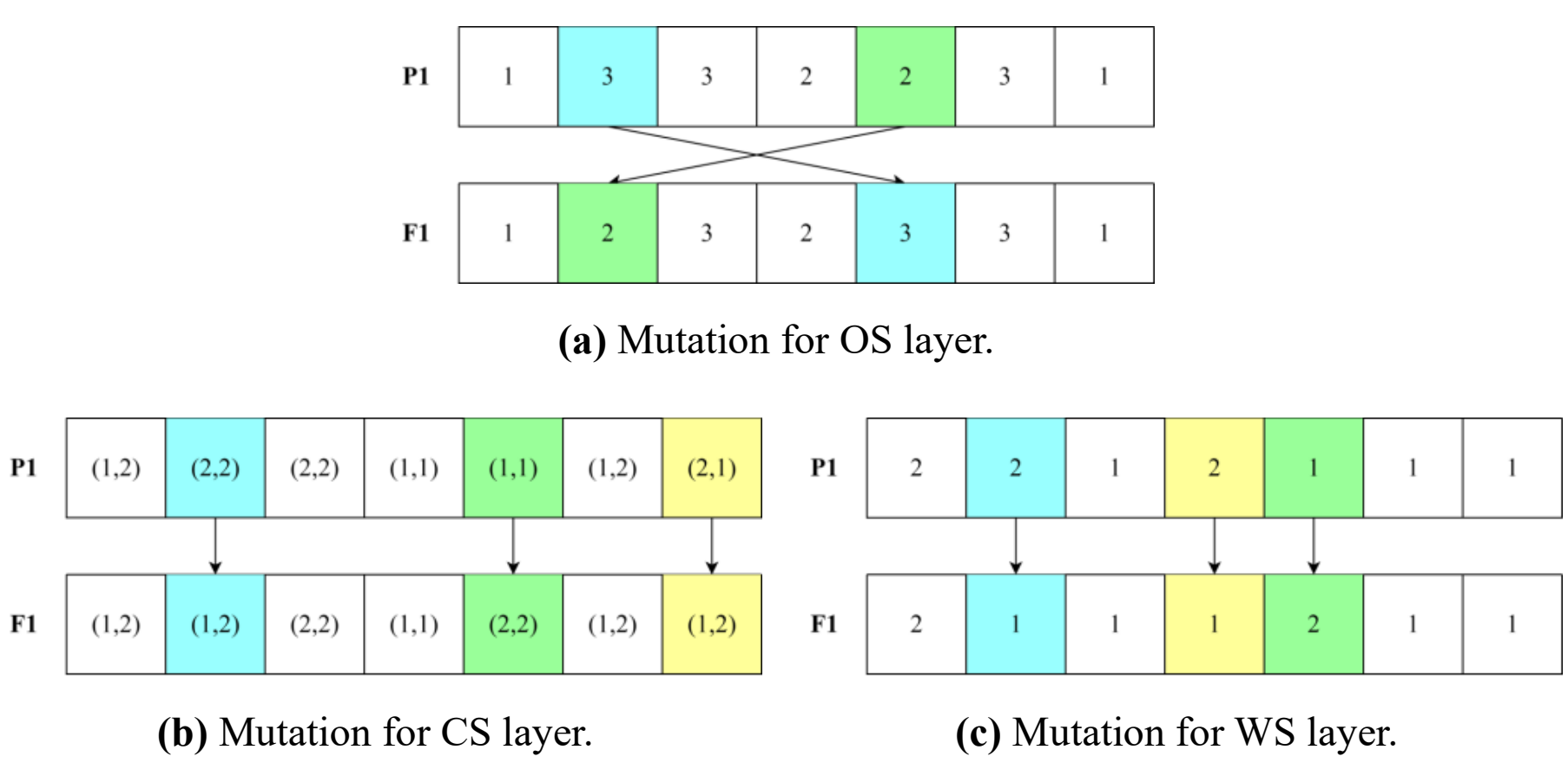

**(a)** Mutation for OS layer.

**(b)** Mutation for CS layer.

**(c)** Mutation for WS layer.

**Fig. 3.** Mutation generator.

# 4 Discussions

To validate the presented mathematical model, we adopt 20 instances [16-18] and extend them for our problem as shown in Table 1. Here, $n$ denotes the number of jobs, $m$ denotes the number of machines, $l$ represents the number of workers, and $c$ represents the number of configurations. The energy consumption values are randomly generated within the range of 3 to 30 units. The number of configurations, $c$, is randomly generated as an integer within the range $[m/4, m/2]$.

The MIP model is implemented using CPLEX 12.6.3 solver, while the MA and GA are executed in Python 3.12.3 and run 20 times. All computations are performed on a personal computer equipped with 12 GB RAM and a Core-i9 CPU.

To compare MA with MIP and GA, the Relative Percentage Difference (RPD) measure is utilized [19]. The RPD is calculated as follows:

$$RPD = \frac{F_2 - F_1}{F_1} \times 100\% \tag{12}$$

where $F_1$ represents the best-known solution obtained by the MA and $F_2$ denotes the best-known solution obtained by other algorithms, such as MIP and GA. The parameters are as follows: population size of 100, crossover rate of 0.8, mutation rate of 0.3, and neighborhood search rate of 0.1.

Fig. 4 illustrates a comparative box plot for the MA and GA over twenty instances. The results presented in Table 1 indicate that the CPLEX yields the best solutions for the first seven instances. However, it becomes impractical for larger instances due to excessive computation times. Although the MIP approach outperforms the MA in terms of solution quality, the solutions provided by the MA are still quite close to those obtained by the MIP. Moreover, the MA has the advantage of significantly shorter running times compared to the MIP. Therefore, the MA proves to be an efficient method for solving smaller instances of the problem.

For the remaining instances (E08 – E20), the MA outperforms other methods in achieving the best solutions. Furthermore, its CPU times are shorter than those of the GA. While the GA is effective in solving the FJSSP, our problem's inclusion of worker and machine configurations expands the solution space, making it challenging to identify the best solution within a limited runtime. The proposed MA is particularly effective due to its three neighborhood search mechanisms, which enhance its capability to escape local optima and find superior solutions.

**Table 1.** Computational results.

| Instance | Size | RPD | | | CPU time (s) | | |
|---|---|---|---|---|---|---|---|
| | $n \times m \times l \times c$ | MA | GA | MIP | MA | GA | MIP |
| E01 | 5×6×3×2 | 4.77 | 10.34 | **0.00** | **63.12** | **63.12** | 91.35 |
| E02 | 5×7×3×2 | 4.94 | 12.46 | **0.00** | **76.55** | 79.12 | 113.97 |
| E03 | 6×7×3×2 | 5.56 | 14.06 | **0.00** | **93.01** | 103.16 | 204.36 |
| E04 | 7×7×3×2 | 7.95 | 9.77 | **0.00** | **121.06** | 154.33 | 801.03 |
| E05 | 7×7×3×3 | 8.32 | 10.31 | **0.00** | **145.87** | 177.89 | 975.74 |
| E06 | 8×7×3×2 | 8.64 | 11.03 | **0.00** | **204.38** | 241.55 | 1964.66 |
| E07 | 8×7×3×3 | 9.55 | 11.79 | **0.00** | **243.47** | 278.00 | 2133.73 |
| E08 | 9×8×4×3 | **0.00** | 20.44 | - | **289.65** | 300.54 | - |
| E09 | 10×6×4×2 | **0.00** | 17.45 | - | **322.02** | 394.04 | - |
| E10 | 10×6×4×3 | **0.00** | 16.30 | - | **360.04** | 421.23 | - |
| E11 | 11×8×4×4 | **0.00** | 17.66 | - | **408.58** | 488.02 | - |
| E12 | 12×8×4×4 | **0.00** | 25.01 | - | **453.97** | 533.33 | - |
| E13 | 15×4×5×2 | **0.00** | 25.48 | - | **675.63** | 702.22 | - |
| E14 | 15×8×6×2 | **0.00** | 25.37 | - | **689.33** | 797.50 | - |
| E15 | 15×8×6×3 | **0.00** | 14.41 | - | **766.09** | 866.69 | - |
| E16 | 10×15×8×4 | **0.00** | 29.39 | - | **705.98** | 846.07 | - |
| E17 | 20×5×4×2 | **0.00** | 19.23 | - | **1358.75** | 1541.55 | - |
| E18 | 20×10×6×3 | **0.00** | 20.07 | - | **1397.43** | 1596.90 | - |
| E19 | 20×10×6×4 | **0.00** | 21.31 | - | **1495.08** | 1788.02 | - |
| E20 | 20×15×8×5 | **0.00** | 20.54 | - | **1683.33** | 1901.25 | - |

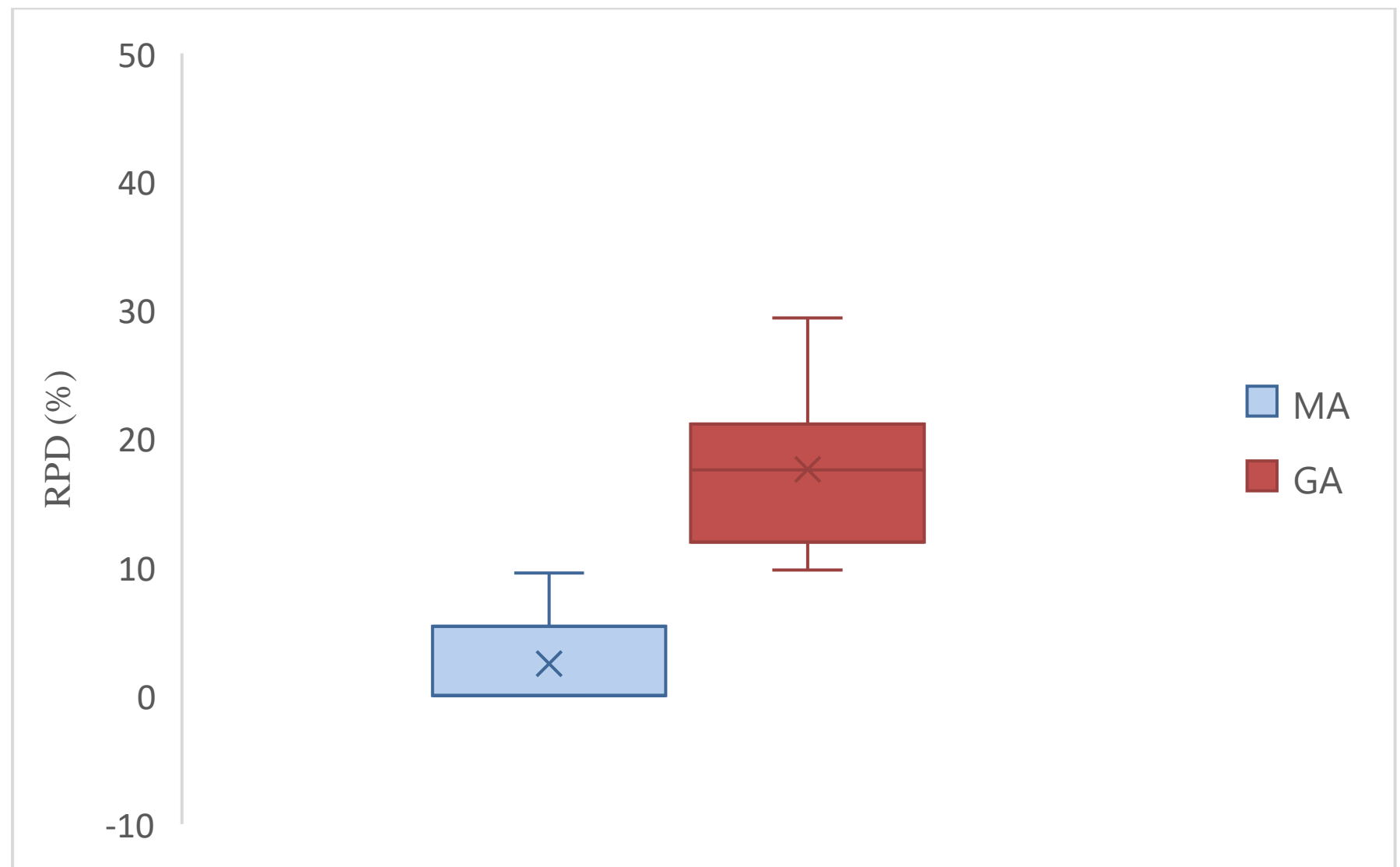

**Fig. 4.** Box plot for the RPD.

## 5 Conclusion

The transition towards Industry 5.0 has necessitated a paradigm shift in manufacturing scheduling approaches. Traditional methods that solely focus on economic objectives are no longer sufficient, as human-centric and environmental considerations have become paramount. The flexible job shop scheduling problem (FJSSP) with reconfigurable machine tools (RMTs) offers a compelling framework to address these multi-faceted priorities. This paper has reviewed integrated FJSSP models that account for human factors like worker skills, ergonomics, and fatigue management, as well as environmental criteria such as energy consumption and emissions. The review has highlighted the scarcity of research efforts that simultaneously optimize these economic, human, and environmental dimensions, especially in the context of reconfigurable machine tools.

The proposed mathematical model contributes by integrating worker assignment constraints, rest periods, and energy consumption objectives into the FJSSP with reconfigurable machines and sequence-dependent setup times. The primary objective of the model is to minimize total energy consumption. Solving large-sized, real-world instances will require the development of efficient meta-heuristic algorithms.

To address this challenge, we developed a memetic algorithm (MA) tailored for this NP-hard combinatorial optimization problem. The key steps of the developed MA include initializing a population, generating offspring using crossover and mutation methods, and applying neighborhood strategies to improve the quality of individuals. This algorithm employs a three-layer chromosome encoding method, including operation sequencing (OS), configuration selection (CS), and worker selection (WS) layers.

The methodology and results of our study demonstrate that the MA provides a competitive alternative to the MIP model, particularly for larger instances, in which exact methods become impractical due to excessive computation times. Although the MIP approach outperforms the MA in terms of solution quality for smaller instances, the MA achieves solutions that are close in quality while significantly reducing computational times. For larger instances, the MA outperforms both the GA and MIP, achieving the best solutions with shorter CPU times. This efficiency is attributed to the three neighborhood search mechanisms integrated into the MA, which enhance its capability to escape local optima and find superior solutions.

Future research can explore multi-objective formulations that capture the trade-offs between economic, human, and environmental criteria, enabling the decision-makers to derive schedules that harmonize these competing objectives. As manufacturing firms navigate the transition to Industry 5.0, holistic scheduling approaches that prioritize human well-being, environmental stewardship, and economic viability will be crucial drivers of operational resilience and sustainable competitiveness. The research directions outlined in this paper pave the way for developing robust scheduling techniques that can unlock the full potential of human-centric, eco-efficient manufacturing in the factory of the future.

**Acknowledgment.** The research has been supported by the European Union under the project ROBOPROX (reg. no. CZ.02.01.01/00/22 008/0004590).

**Disclosure of Interests.** The authors have no competing interests to declare that are relevant to the content of this article.